\documentclass[final]{cvpr}

\usepackage{times}
\usepackage{epsfig}
\usepackage{graphicx}
\usepackage{subfigure}
\usepackage{rotating}
\usepackage{amsmath}
\usepackage{amssymb}
\usepackage{bbm}
\usepackage{mathtools}
\usepackage{url}
\usepackage{color}
\usepackage{float}
\usepackage{multirow}

\usepackage[pagebackref=true,breaklinks=true,colorlinks,bookmarks=false]{hyperref}

\def\cvprPaperID{10966} 
\def\confYear{CVPR 2021}

\begin{document}
\title{Frequency-aware Discriminative Feature Learning Supervised by Single-Center Loss for Face Forgery Detection }

\author{Jiaming Li$^{1}$\qquad Hongtao Xie$^{1}$\thanks{Corresponding author}  \qquad Jiahong Li$^{2}$ \qquad Zhongyuan Wang$^{2}$ \qquad Yongdong Zhang$^{1}$\\
$^{1}$University of Science and Technology of China\qquad $^{2}$Kuaishou Technology \\
{\tt\small ljmd@mail.ustc.edu.cn\qquad\{htxie,zhyd73\}@ustc.edu.cn\qquad\{lijiahong,wangzhongyuan\}@kuaishou.com}
}

\maketitle

\begin{abstract}
   Face forgery detection is raising ever-increasing interest in computer vision since facial manipulation technologies cause serious worries.
   Though recent works have reached sound achievements, there are still unignorable problems: 
   a) learned features supervised by softmax loss are separable but not discriminative enough, since softmax loss does not explicitly encourage intra-class compactness and inter-class separability;
   and b) fixed filter banks and hand-crafted features are insufficient to capture forgery patterns of frequency from diverse inputs. 
   To compensate for such limitations, a novel frequency-aware discriminative feature learning framework is proposed in this paper. 
   Specifically, we design a novel single-center loss (SCL) that only compresses intra-class variations of natural faces while boosting inter-class differences in the embedding space.
   In such a case, the network can learn more discriminative features with less optimization difficulty.
   Besides, an adaptive frequency feature generation module is developed to mine frequency clues in a completely data-driven fashion. 
   With the above two modules, the whole framework can learn more discriminative features in an end-to-end manner. 
   Extensive experiments demonstrate the effectiveness and superiority of our framework on three versions of the FF++ dataset.
\end{abstract}
\section{Introduction}
Benefiting from the great progress made in deep learning, the Variational AutoEncoders~\cite{kingma2013auto,rezende2014stochastic} and Generative Adversarial Networks based~\cite{goodfellow2014generative}
face manipulation technology~\cite{shen2020interpreting,lee2020maskgan,wu2020cascade} enables ordinary people without professional skills and equipment to generate high-quality forged faces.
Derived from that, certain free apps~\cite{FaceApp} and open-source projects~\cite{Deepfake,FaceSwap01} quickly arise and gain popularity explosively. 
Unluckily, the technology may be abused for malicious purposes, causing severe trust issues in our society.  
Although digital forensics experts can analyze some influential videos for evidence of manipulation, 
they will be helpless in reviewing countless videos uploaded to the Internet every day. 
Thus, it is of high significance to develop efficient automatic detection algorithms.
\begin{figure} 
   \begin{center}
   \includegraphics*[width=0.8\linewidth]{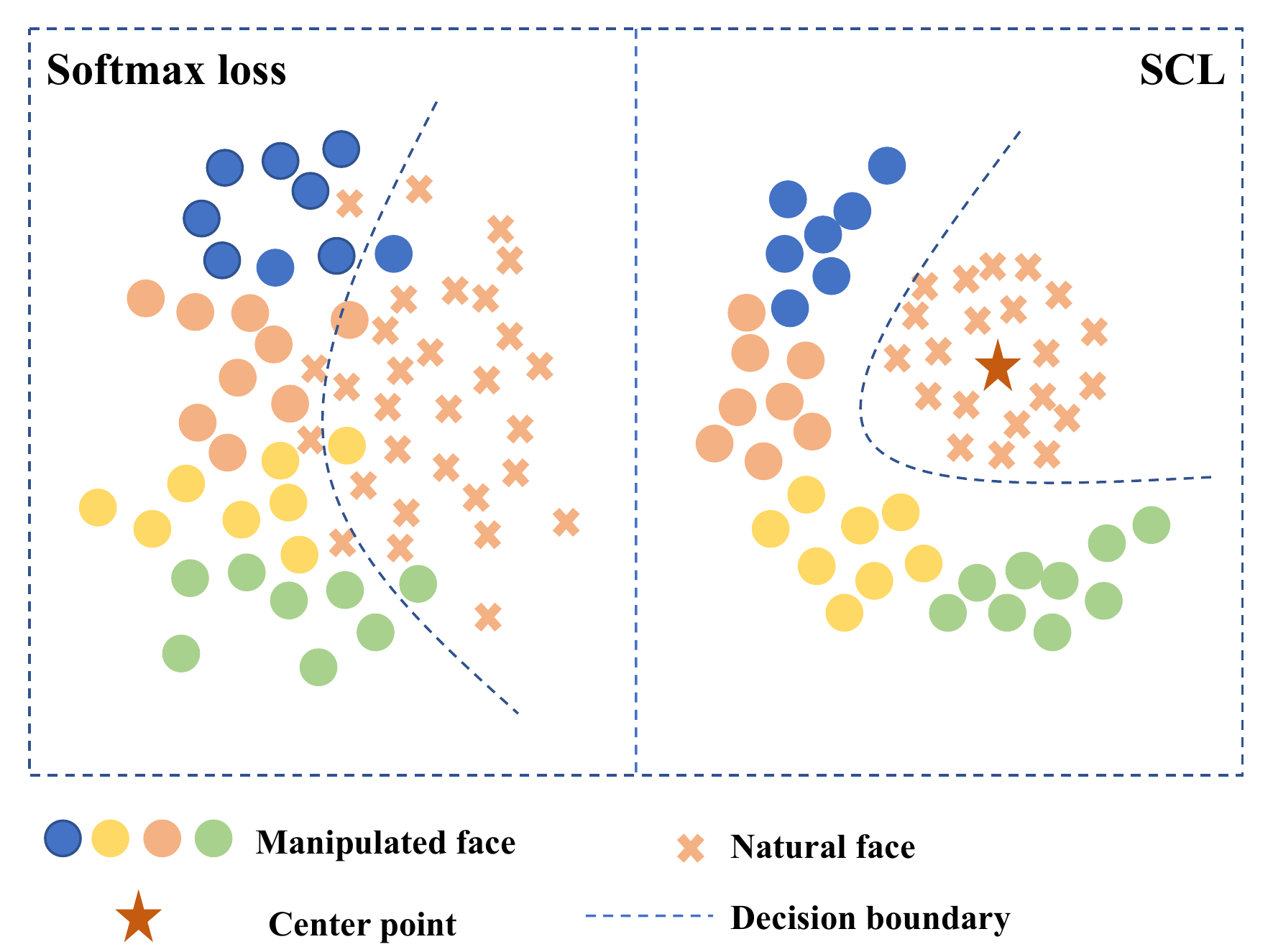}
   \end{center}
      \caption{The feature distribution of samples in the embedding space. Left: learned features supervised by softmax loss are 
      broadly separable but not discriminative enough, since the intra-class compactness and inter-class separability are not explicitly constrained.
      Right: our SCL only encourages the intra-class compactness of natural faces when constraining inter-class separability.}
   \label{fig:idea}
   \vspace{-0.4cm} 
\end{figure}

Towards such a concern, many methods have been proposed successively. 
Early research is keen on utilizing hand-crafted features or modifying the structure of existing 
neural networks\cite{yang2019exposing,agarwal2019protecting,afchar2018mesonet,rahmouni2017distinguishing}.
However, with remarkable progress made in facial synthesis technology~\cite{karras2019style,yang2020one,deng2020disentangled}, such methods have been unable to reliably detect face forgery.
After that, the research mainstream is gradually turning to methods that introduce different information and prior knowledge into backbone networks
~\cite{dang2020detection,qi2020deeprhythm,masi2020two}.
For example, DeepRhythm~\cite{qi2020deeprhythm} utilizes the minuscule periodic changes of skin color due to blood pumping through the face. 

In essence, all current popular detection methods are using the powerful data fitting capability of neural networks 
to extract discriminative features for face forgery detection.
And detection methods based on deep learning usually pose face forgery detection as a binary classification problem 
and use softmax loss\footnote{Following~\cite{liu2016large}, we define the softmax loss as the combination of the
last fully connected layer, softmax function, and cross-entropy loss.} to supervise the training of CNN networks. 
However, learned features supervised by softmax loss are not discriminative enough, 
since softmax loss does not explicitly encourage intra-class compactness and inter-class separability, as illustrated in the left of Figure~\ref{fig:idea}.
Recent work~\cite{kumar2020detecting} has noticed this problem and attempted to utilize triplet loss~\cite{schroff2015facenet} to extract discriminative features.
However, regular metric learning methods usually indiscriminately encourage the intra-class compactness of natural and manipulated faces in the embedding space. 
Additionally, feature distributions of manipulated faces vary from one manipulation method to another considering various GAN fingerprints~\cite{yu2019attributing} and some unique operations, as shown in the left of Figure~\ref{fig:idea},
making it nontrivial to aggregate all the manipulated faces. Therefore, constraining intra-class compactness of samples generated by varied manipulation methods
usually leads to a sub-optimal solution because of optimization difficulty and even damages the performance owing to overfitting.

In addition, frequency-related cues are increasingly important for forgery detection since it's hard to find visual forgery clues.
Although some studies~\cite{masi2020two,cozzolino2018forensictransfer,wang2020cnn,durall2019unmasking} have introduced frequency information and achieved remarkable results,
their abilities to extract discriminative features are limited because of employing fixed filter banks and hand-crafted features.
These methods based on incomprehensive prior knowledge are insufficient to capture subtle forgery patterns from the frequency domain due to the diversity of background, gender, age, manipulation methods, $\etc$

\begin{figure*}
   \begin{center}
      \includegraphics*[width=0.8\linewidth]{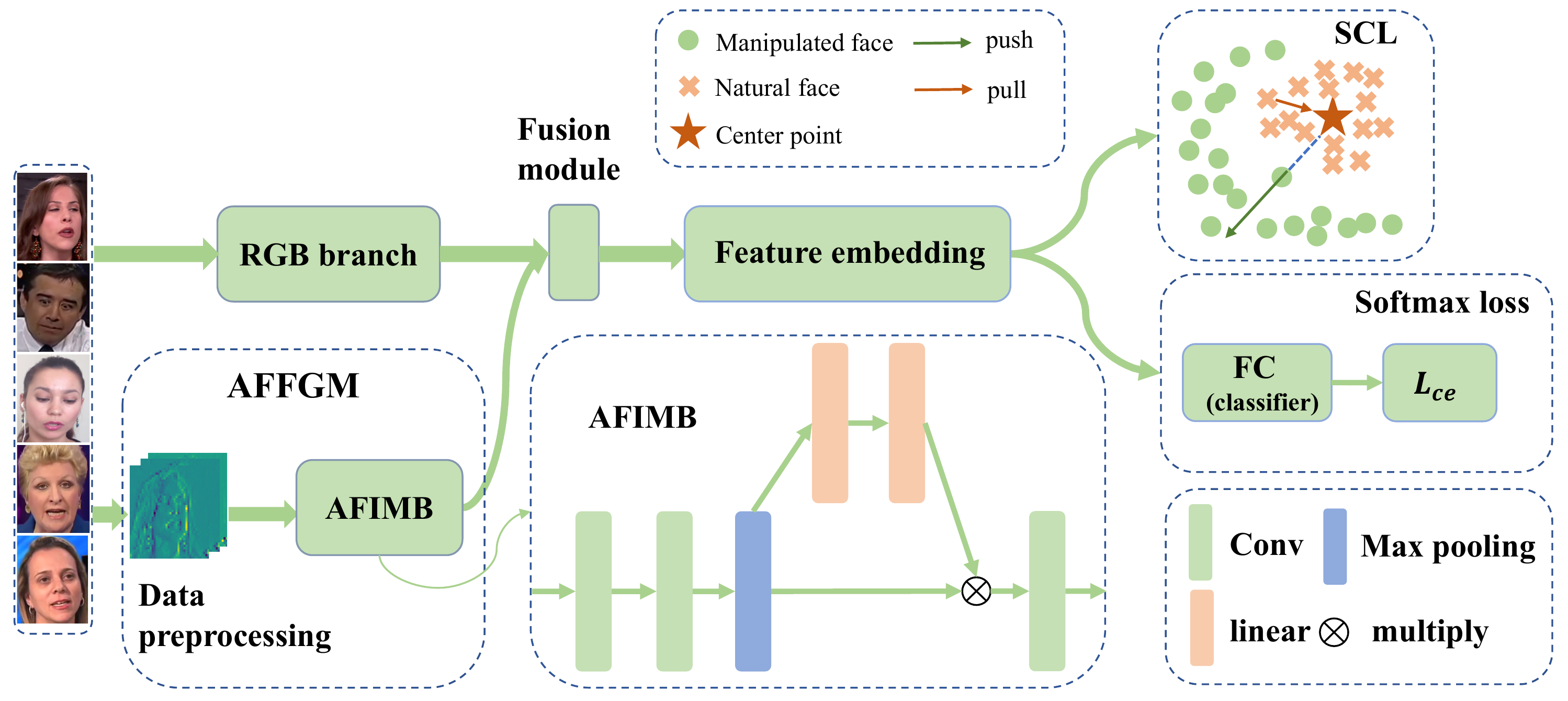}
   \end{center}
      \caption{The Frequency-aware Discriminative Features Learning framework. AFFGM stands for the adaptive frequency feature generation module. 
      AFIMB represents the adaptive frequency information mining block. FC represents the fully connected layer and $L_{ce}$ represents the cross-entropy loss. The whole framework is trained end-to-end under the joint supervision of SCL and softmax loss.}
   \label{fig:piple}
   \vspace{-0.4cm}
   \end{figure*}

With the above thoughts in mind, we propose a novel Frequency-aware Discriminative Feature Learning framework(FDFL).
Explicitly, our framework mainly addresses two problems: a) how to adopt metric learning to learn more discriminative features for face forgery detection; and b) how to 
adaptively extract frequency-related features.
Corresponding to the two problems, two sub-modules are developed: single-center loss (SCL) and adaptive frequency feature generation module (AFFGM), as shown in Figure~\ref{fig:piple}.
In specific, our single-center loss aims at only reducing intra-class variations of natural faces while increasing inter-class differences in the embedding space,
as shown in the right of Figure~\ref{fig:idea}.
To this end, SCL minimizes the distance from representations of natural faces to the center point. 
Meanwhile, SCL encourages the distance from manipulated faces to the center point greater than from natural faces by at least a margin. 
Unlike regular metric learning methods, SCL does not restrict the intra-class compactness of manipulated faces, which agrees better with the characteristics of feature distribution of manipulated faces.
Therefore, the network supervised by SCL can learn more discriminative features with less optimization difficulty. 
As for frequency-related features, we develop an AFFGM consisting of a special data preprocessing 
and adaptive frequency information mining block (AFIMB). The data preprocessing keeps the position relationship of image blocks in the spatial domain consistent with their position relationship in the frequency domain. 
In such a case, the preprocessed data is able to directly employ the existing convolution network. The AFIMB adaptively mines frequency clues in a data-driven fashion, which 
avoids utilizing too much incomprehensive prior knowledge. Compared to fixed filter banks and hand-crafted features, 
AFFGM can capture forgery clues more flexibly in the frequency domain.

Extensive experiments demonstrate the effectiveness and superiority of our framework and we achieve
state-of-the-art results on three versions of the FF++ dataset~\cite{rossler2019faceforensics++}. Our contributions can be summarized as follows:
\begin{itemize}
   \setlength{\itemsep}{0pt}
   \setlength{\parsep}{0pt}
   \setlength{\parskip}{0pt}
   \item We propose a novel Frequency-aware discriminative feature learning framework which adopts metric learning and adaptive frequency features learning for face forgery detection.
   \item A single-center loss is designed to only compress intra-class variations of natural faces while boosting inter-class differences in the embedding space. 
   \item An adaptive frequency feature generation module is developed to mine subtle artifacts from the frequency domain in a data-driven fashion.
\end{itemize}
\section{Related work}
With the development of neural networks and computer graphics, 
a new generation of face manipulation technology based on the Variational AutoEncoders~\cite{kingma2013auto,rezende2014stochastic} and Generative Adversarial Networks~\cite{goodfellow2014generative} has been widely used. 
Correspondingly, face forgery detection has gradually become a research hotspot. 
In this section, we will briefly review previous works.
\paragraph{Face forgery detection}
Early works focus on utilizing hand-crafted features or modifying the structure of existing 
neural networks~\cite{yang2019exposing,agarwal2019protecting,jung2020deepvision, afchar2018mesonet,rahmouni2017distinguishing} to detect face forgery. 
Yang $\etal$~\cite{yang2019exposing} utilize the inconsistency of the head pose estimated from the central face and the whole face to identify manipulated faces.
MesoNet~\cite{afchar2018mesonet} designs a shallow neural network that consists of two inception modules and two classic convolution layers.
Though sound performances were achieved at that time, those methods are incapable of reliably detecting face forgery now due to the rapid development of face forgery technology.
Especially when powerful general feature extractors like xception~\cite{chollet2017xception} are applied to forgery detection, the performance of early works is even more unsatisfactory.
Therefore, the research mainstream is gradually turning to approaches which introduce different information and prior knowledge into the backbone network to detect face forgery
~\cite{dang2020detection,qi2020deeprhythm,masi2020two}.
Dang $\etal$~\cite{dang2020detection} introduce location information of manipulated regions to guide the network to focus on key regions.
Qi $\etal$~\cite{qi2020deeprhythm} exploit bioinformatics that skin color will present minuscule changes periodically due to blood pumping through the face.
Face X-ray~\cite{li2020face} innovatively uses self-generated data to train the network to locate blending boundaries, which greatly improves the generalization ability.
Two-branch~\cite{masi2020two} utilizes fixed filter banks to extract frequency information, which limits the ability to extract discriminative features.
In our work, we exploit a simple and effective module to adaptively mine frequency clues. 
\paragraph{Metric learning}
Although metric learning has shown its advantages in face recognition~\cite{schroff2015facenet} and person re-identification (re-ID)~\cite{hermans2017defense}, 
learning discriminative features with deep metric learning for face forgery detection is more or less neglected. Center loss~\cite{wen2016discriminative} and triplet loss~\cite{schroff2015facenet}
are the two most relevant metric learning methods to our work.
Center loss~\cite{wen2016discriminative} is designed to learn a center for features of each class and drive features of the same class closer to their corresponding center.
Obviously, one disadvantage of center loss is that it ignores inter-class separability.
Triplet loss~\cite{schroff2015facenet} encourages features of data points with the same identity to get closer than those with different identities.
However, triplet loss may suffer from the problem of time-consuming mining of hard triplets and dramatic data expansion.
Kumar $\etal$~\cite{kumar2020detecting} utilizes the network with the supervision of triplet loss to detect face forgery. 
But triplet loss performs poorly on the imagenet pre-trained backbone. Two-branch~\cite{masi2020two} proposes a novel loss which compresses the variability of natural faces and pushes away the manipulated faces. 
But its motivation comes from anomaly detection and the approach is very different from our SCL in many aspects.
For example, our center point is updatable, while the center point of the two-branch is fixed. Additionally,
two-branch constrains the absolute distance from all samples to the center point, 
whereas our SCL constrains the relative distance between natural and manipulated samples to the center point.
\section{Proposed method}
\subsection{Overview}\label{sec:overview}
Aiming at solving the problems of previous methods in discriminative feature learning and frequency information mining, we propose a frequency-aware discriminative feature learning framework. 
As illustrated in Figure~\ref{fig:piple}, our framework extracts features from the RGB domain and frequency domain at the same time
and merges them in the early stage of the entire framework. After going through a feature embedding, high-level representations are obtained.
At the end of the framework is a classifier that outputs the prediction results of input samples.
The mining of frequency clues is achieved by our AFFGM (see Sec.~\ref{sec:AFFGM}).  
We fuse the frequency domain features and RGB domain features
with a simple point-wise convolution block,
which contributes to the reduction of parameters and computational expense.
Finally, with the joint supervision of our single-center loss (see Sec.~\ref{sec:SCL}) and softmax loss, the network learns an embedding space where 
natural faces are clustered around the center point, while manipulated ones are far away from the center point.
\subsection{Adaptive frequency features generation module}\label{sec:AFFGM}
\begin{figure}
   \begin{center}
      \includegraphics*[width=0.8\linewidth]{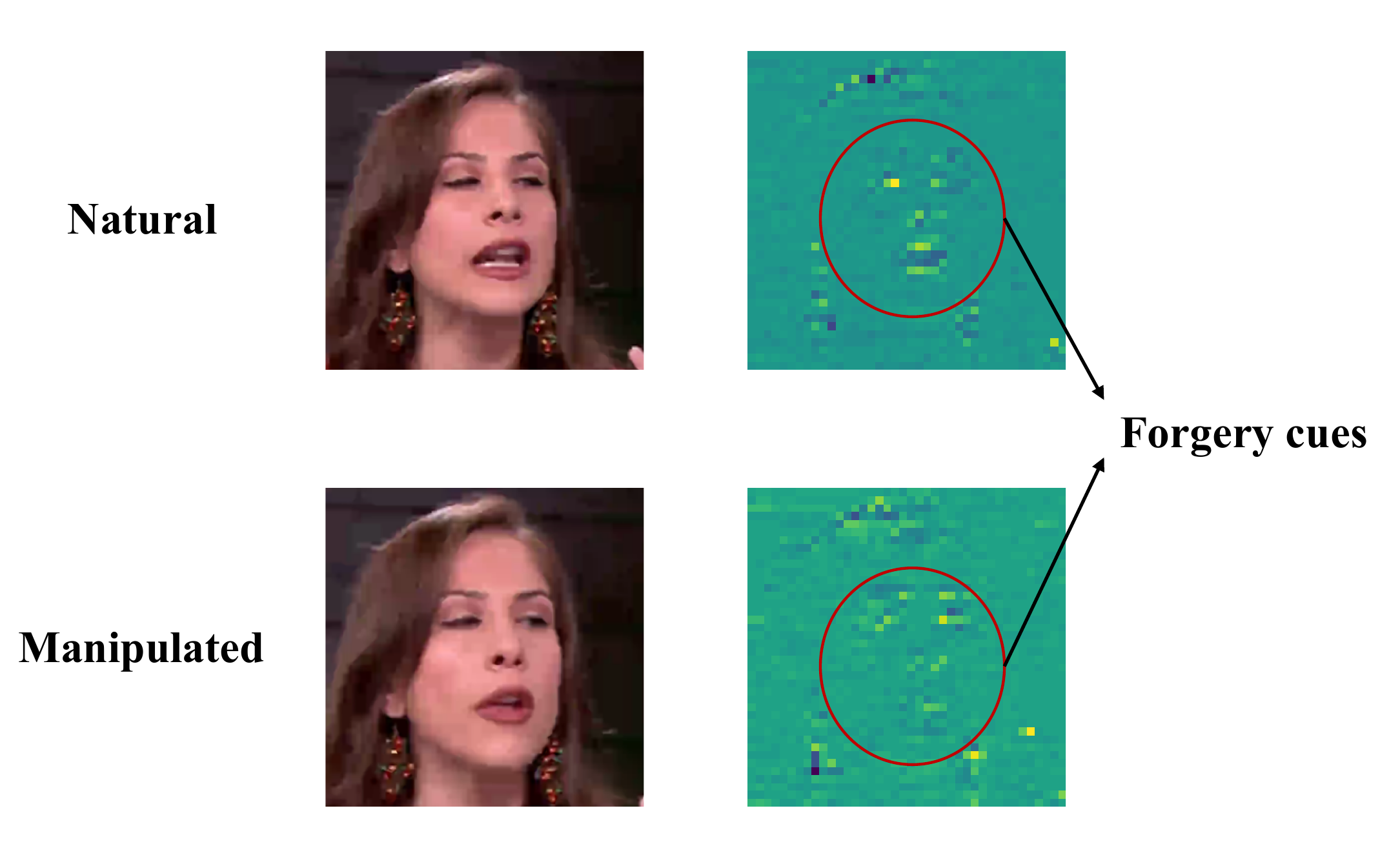}
   \end{center}
      \caption{Inconsistency in the frequency domain could serve as an important forgery cue. The visualization of energy distribution in a certain frequency band is shown in the right column.}
   \label{fig:dct}
   \vspace{-0.4cm}
   \end{figure}

\begin{figure}
   \begin{center}
      \includegraphics*[width=0.8\linewidth]{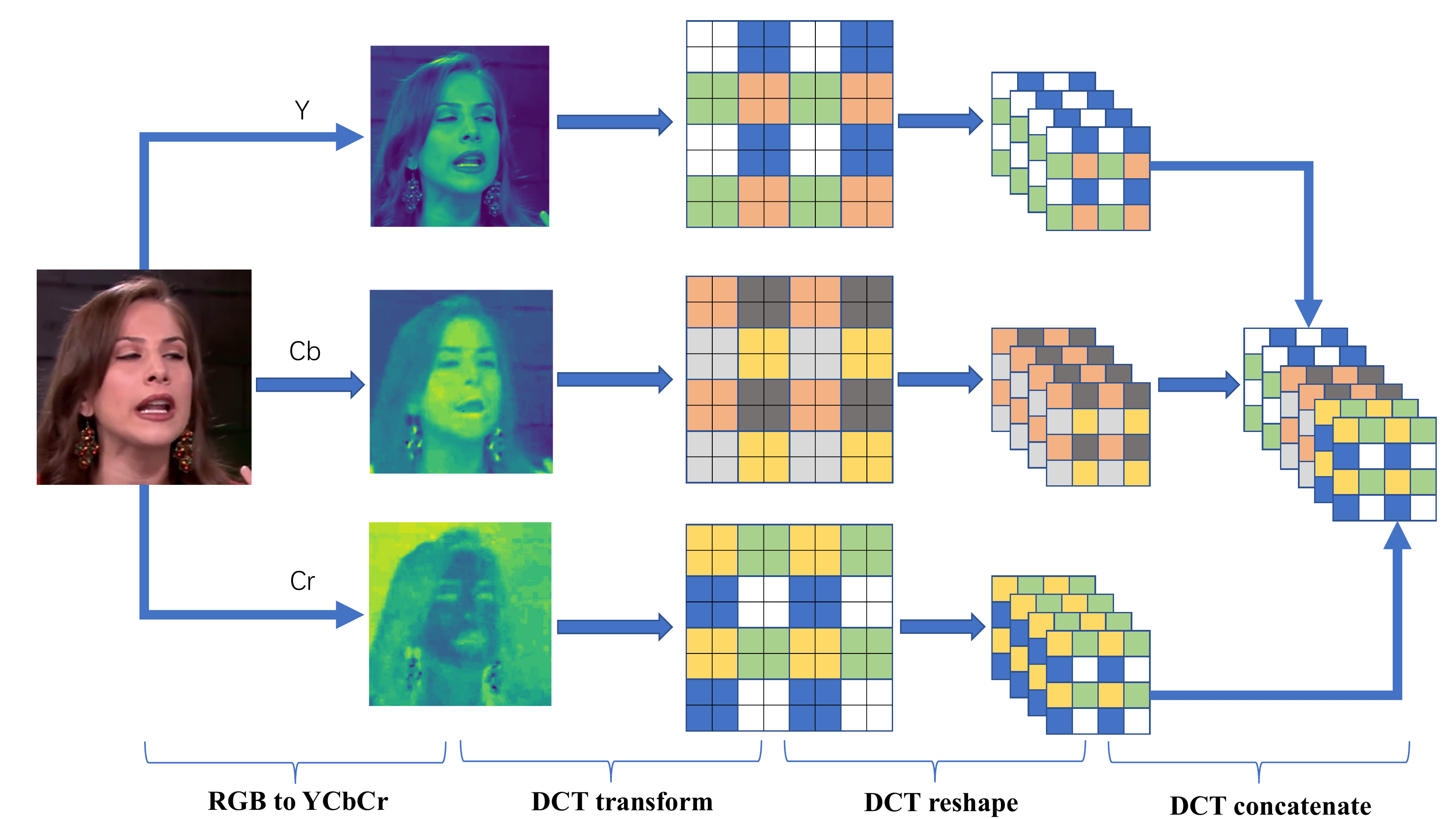}
   \end{center}
      \caption{The pipeline of data preprocessing of AFFGM.}
   \label{fig:data}
   \vspace{-0.4cm}
   \end{figure}

   With the success in synthesizing realistic faces, it's harder to find visual forgery clues.
   But the discrepancy between natural and manipulated faces in the frequency domain, especially in middle and high frequency bands, is pretty apparent as illustrated in Figure~\ref{fig:dct}.
   Previous studies mostly use fixed filter banks or hand-crafted methods from other fields to extract frequency information~\cite{cozzolino2018forensictransfer,durall2019unmasking,wang2020cnn, masi2020two}. 
   However, considering the diversity of background, gender, age, skin color, and especially manipulation methods, these methods based on incomprehensive prior knowledge are insufficient to capture forgery patterns of frequency.  
   In order to tackle this problem, inspired by \cite{gueguen2018faster,Xu_2020_CVPR,min2020multi,liu2019bidirectional}, 
   we propose an adaptive frequency feature generation module (AFFGM) to efficiently mine subtle artifacts from the frequency domain. 
   Our AFFGM consists of two parts: data preprocessing and adaptive frequency information mining block. 
   Next, we will introduce them respectively.
\paragraph{Data preprocessing}
The pipeline of data preprocessing is shown in Figure~\ref{fig:data}. 
First, input RGB images are transformed into YCbCr color space.
Next, the 2D DCT transformation is applied to each $8\times8$ block of images.
It's worth noting that the two steps above are also widely used in current popular image compression standards, $\eg$, JPEG. We think that will contribute to forgery detection from two aspects. 
On the one hand, the acceleration tools of existing compression algorithms 
can help improve the computational efficiency of our preprocessing. On the other hand, it will make our method more compatible with traces caused by compression.
After that, the DCT-transformed coefficients from the same frequency band in all $8\times 8$ blocks are grouped into a channel with their original position relationship retained.
Therefore, the transformed images can directly exploit existing neural networks. 
Finally, all frequency channels are concatenated together to form one tensor.
The shape of input images will change before and after preprocessing.
Suppose the shape of the original input image is $H\times W\times 3$, then the shape of the input tensor becomes $H/8\times W/8\times 192$ after data preprocessing.
Moreover, most energy of transformed images is concentrated on the low-frequency bands while the middle-frequency and high-frequency bands play more significant roles in forgery detection. 
Therefore, every frequency channel is normalized by the mean and variance calculated from the training dataset.
\paragraph{Adaptive frequency information mining block}
Unlike previous methods, our AFFGM learns the frequency feature in a data-driven way, which avoids overly depending on incomprehensive prior knowledge.
As illustrated in Figure~\ref{fig:piple}, we empirically design a simple and effective network block to extract frequency features. 
In specific, the preprocessed data first passes through a layer of $3\times3$ convolution block with three groups\cite{ioannou2017deep}. 
That means the data from different channels of Y, Cb, Cr is processed separately. Then, it goes through an ordinary $3\times 3$ convolution block and a max-pooling layer successively.
In that process, information from different channels of Y, Cb, Cr interacts with each other. After that, we employ a channel attention block which consists of the aforementioned max-pooling layer and two linear layers for the sake of
feature enhancement. Finally, an ordinary $1\times 1$ convolution layer is used to further extract frequency-related features.
\subsection{Single-center loss}\label{sec:SCL}
   Current face forgery detection methods based on deep learning usually use softmax loss to supervise network training. 
   However, the learned features supervised by softmax loss are essentially not discriminative enough, 
   since softmax loss only focuses on finding a decision boundary to separate different classes.
   The intra-class compactness and inter-class separability are not explicitly considered.
   Obviously, deep metric learning is a promising solution. However, most metric learning methods, such as triplet loss~\cite{schroff2015facenet} and center loss~\cite{wen2016discriminative},
   usually indiscriminately compress intra-class variations of natural and manipulated faces in embedding space. 
   While feature distributions of manipulated faces vary from one manipulation method to another.
   That is because GAN fingerprints~\cite{yu2019attributing}, manipulated region, and other unique operations, $\eg$, post-processing techniques, lead to specific artifacts
   for each manipulation method.
   For example, Deepfakes~\cite{Deepfake} generates the whole face while NeuralTextures~\cite{thies2019deferred}
   only manipulates the mouth region of the target person. Intuitively, their distribution in the embedding space should be evidently different.
   A side evidence would be that the generalization ability of features learned by supervised learning is significantly weakened on unseen manipulation methods. 
   This implies that features learned by supervised learning are highly related to manipulation methods.
   The feature differences of samples generated by different manipulation methods make it difficult to aggregate all the manipulated faces. 
   Therefore, indiscriminately constraining intra-class compactness in embedding space usually leads to a sub-optimal solution due to optimization difficulty and even damages the performance owing to overfitting. 
   In order to solve this problem, we devise a novel single-center loss.
\paragraph{Definition}
   As Figure~\ref{fig:piple} indicates that the goal of SCL is to minimize the distance from the representations of natural faces to the center point 
   and to simultaneously push the representations of manipulated ones away from the center point.
   Let the given training dataset ${(x^{i}, y^{i})}^{N}_{i=1}$ consist of N samples $x_{i} \in  X$ with the associated labels $y_{i} \in \{0,1\}$.
   And these samples are embedded into D-dimensional vectors with a neural network denoted by $f_{\theta}(\cdot)$. In our SCL, 
   we just set the center point $C$ of natural faces. For simplicity, we adopt $f_{i}$ to represent $f(x^{i})$ in the following paper.
   Similar to center loss, our method updates the parametric centers $C$ at each iteration based on a mini-batch.
   Given a batch of training data, we define SCL as: 
   \begin{equation} 
         L_{sc} =M_{nat}+ max(M_{nat}-M_{man}+m\sqrt{D},0) \label{eq:SCL}
   \end{equation}
   where $M_{nat}$ represents the mean Euclidean distance between representations of natural faces and the center point C in a batch. 
   And $M_{man}$ represents the mean Euclidean distance between representations of manipulated faces and center point C. Their functions are denoted as: 
   \begin{equation}
      M_{nat}=\frac{1}{|\Omega_{nat}|}\sum_{i \in \Omega_{nat}}{\parallel f_{i}-C\parallel_{2}}
   \end{equation}
   \begin{equation}
      M_{man}=\frac{1}{|\Omega_{man}|}\sum_{i \in \Omega_{man}}{\parallel f_{i}-C\parallel_{2}}
   \end{equation}
   where $\Omega_{nat}$ and $\Omega_{man}$ represent the representation sets of natural faces and manipulated faces respectively.
   As Eq.~\eqref{eq:SCL} shows, our SCL makes representations of natural faces aggregated around the center point. And it also pushes the distance from representations of manipulated faces to the center point greater 
   than from natural faces by a margin. The Euclidean distance we employ is related to the arithmetic square root of feature dimension,
   and hence in order to set the hyperparameter easily, the margin is designed as m$\sqrt{D}$.

   To compute the back-propagation gradients of the input feature embeddings and the center point, 
   we assume there are $s$ natural faces and $t$  manipulated faces in a batch. 
   And $y_{i}=0$ and $y_{i}=1$ represent i-th sample is a natural face and manipulated face respectively.
   The $\mathbbm{1}[condition]$ is an indicator function which outputs 1 if the condition is satisfied and outputs 0 otherwise.
   For simplicity, we define $$L=M_{nat}-M_{man}+m\sqrt{D}.$$
   Then the derivatives of our SCL loss Eq.~\eqref{eq:SCL} with respect to
   the feature embedding of i-th sample $\frac{\partial L_{sc}}{\partial f_{i}}$ and center point $\frac{\partial L_{sc}}{\partial C}$
   can be calculated as follows:
   \begin{equation}
         \frac{\partial L_{sc}}{\partial f_{i}} = \begin{dcases}
            \frac{f_{i}-C}{s\cdot||f_{i}-C||_{2}}\cdot(1+\mathbbm{1}[L>0]), &y_{i}=0;\\
            -\frac{f_{i}-C}{t\cdot||f_{i}-C||_{2}} \cdot\mathbbm{1}[L>0], &y_{i}=1.
         \end{dcases}    
   \end{equation}
   \begin{equation}
      \begin{split}
      \frac{\partial L_{sc}}{\partial C} = 
         &-\frac{1}{s}(\sum_{i \in \Omega_{nat}}\frac{f_{i}-C}{||f_{i}-C||_{2}})\cdot(1+\mathbbm{1}[L>0])\\
         &+\frac{1}{t}(\sum_{i \in \Omega_{man}}\frac{f_{i}-C}{||f_{i}-C||_{2}})\cdot\mathbbm{1}[L>0].
      \end{split}
   \end{equation}
   The parametric center of SCL is randomly initialized and updated based on the mini-batches instead of the whole datasets, which will cause unstable training.
   Therefore, we introduce softmax loss with global information to guide the update of the center point.
   Moreover, softmax loss focuses on mapping the samples to discrete labels and our SCL aims to apply metric learning to the learned embeddings directly. 
   Combining the two losses is beneficial to achieve more discriminative embeddings. The total loss can be written as:
   \begin{equation}
      L_{total} = L_{softmax}+\lambda L_{sc}
      \label{eq:loss}
   \end{equation}
   where $\lambda$ is a hyper-parameter which controls the trade-off between the SCL and softmax loss. 
\section{Experiments}
In this section, we first introduce the overall experiment setup and then present extensive experimental results
to demonstrate the effectiveness and superiority of our approach.
\subsection{Experimental setup}
\paragraph{Dataset}
In order to facilitate comparison, our experiments are conducted on the FF++~\cite{rossler2019faceforensics++} dataset.
FF++ is a large-scale video dataset consisting of 1000 original videos that have been manipulated 
by four face manipulation methods:
DeepFakes \cite{Deepfake}, Face2Face \cite{thies2016face2face}, FaceSwap\cite{FaceSwap01}, and NeuralTextures \cite{thies2019deferred}.
According to various compression factors, there are three versions of FF++ dataset: c0 (raw), c23 (light compression), and c40 (heavy compression).
Our experiments are mainly conducted on the c40 version, the most challenging case.
As for dataset preprocessing, we sample 20 frames from each manipulated video and 80 frames from each original video.
Compared to the setting of \cite{rossler2019faceforensics++}, the number of frames we use for training is pretty less.
Besides, we utilize retinaface\cite{deng2020retinaface} to detect faces in each frame. 

\paragraph{Evaluation metrics}
Following \cite{masi2020two}, we report video-level AUC score and pAUC~\cite{mcclish1989analyzing} score
by respectively averaging the AUC scores and pAUC scores of each frame in a video.
pAUC is a global metric at a low false alarm rate. Given the significant class imbalance in the real world, 
pAUC can better reflect the performance of methods in the real world.
To facilitate comparison with other methods, we also report the accuracy score.
Besides, some visualizations (t-SNE~\cite{maaten2008visualizing}) are also reported to further evaluate the performance.

\paragraph{Implementation detail}
Our framework is implemented by PyTorch~\cite{paszke2019pytorch}. We use xception~\cite{chollet2017xception} pre-trained on imagenet, including the final fully connected layer, as our RGB branch and feature embedding
of FDFL, which means D in Eq.~\eqref{eq:SCL} is equal to 1000.
The fusion module is inserted between the entry flow and the middle flow of xception and the face forgery classifier is a simple FC layer with two nodes.
The proposed modules, including the center point of SCL, are all initialized randomly. 
More details and hyper-parameters are provided in the supplementary material.
\subsection{Ablation study}
We perform the ablation study to analyze the effects of each component in FDFL, especially our SCL. 
All experiments are conducted on the challenging c40 version of the FF++ dataset. 
\subsubsection{SCL}
In this section, we will show the relevant results of SCL experiments in detail to validate the effectiveness and superiority of SCL.
We conduct all experiments including triplet loss and center loss only based on xception~\cite{chollet2017xception}.
\paragraph{Parameter influence}
As indicated by the loss function in Eq.~\eqref{eq:loss}, the margin m and the weight $\lambda$ may affect the final combination of the losses.
Specifically, $\lambda$ in Eq.~\eqref{eq:loss} controls the trade-off between softmax loss and SCL loss. 
And m controls the relative distance between natural and manipulated faces to the center point in the embedding space.
To study the impact of the two hyper-parameters, we present an empirical analysis on the c40 version of the FF++ dataset.

\begin{figure}
      \subfigure[]{
         \begin{minipage}[t]{0.45\linewidth}
            \label{fig:para_1}  
         \centering
          \includegraphics[width=1.\linewidth]{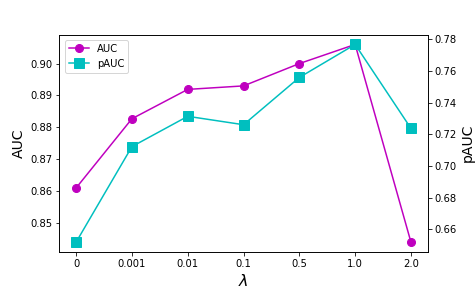}
         \end{minipage}
      }
      \subfigure[]{
         \begin{minipage}[t]{0.45\linewidth}
            \label{fig:para_2}
         \centering
          \includegraphics[width=1.\linewidth]{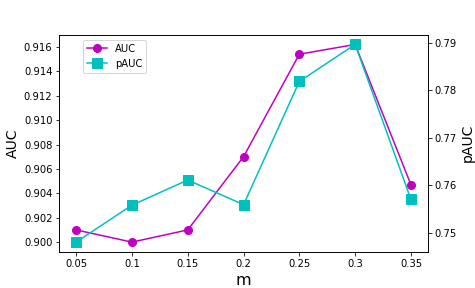}
         \end{minipage}
      }
   \caption{The detection performances achieve by (a) varying $\lambda$ when m is fixed as 0.1 and (b) varying m when $\lambda$ is fixed as 0.5.}  
   \vspace{-0.4cm} 
\end{figure}

\begin{figure*}
   \begin{center}
   \subfigure[]{
      \begin{minipage}[t]{0.2\linewidth}
      \centering
       \includegraphics[width=0.8\linewidth]{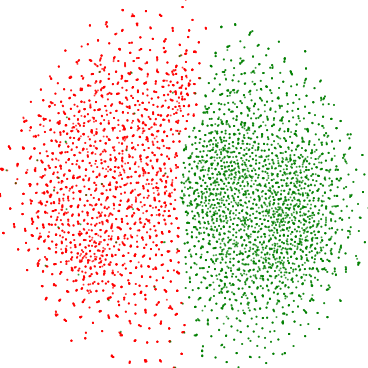}
      \end{minipage}
   }
   \subfigure[]{
      \begin{minipage}[t]{0.2\linewidth}
      \centering
       \includegraphics[width=0.8\linewidth]{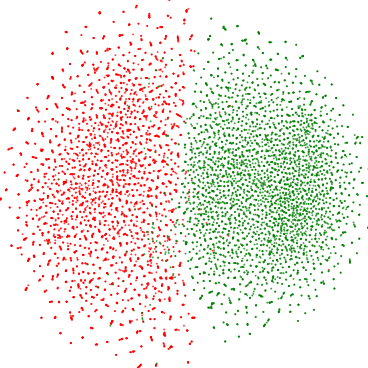}
      \end{minipage}
   }
   \subfigure[]{
      \begin{minipage}[t]{0.2\linewidth}
      \centering
       \includegraphics[width=0.8\linewidth]{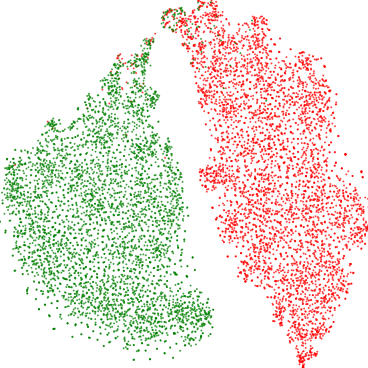}
      \end{minipage}
   }
   \subfigure[]{
      \begin{minipage}[t]{0.2\linewidth}
      \centering
       \includegraphics[width=0.8\linewidth]{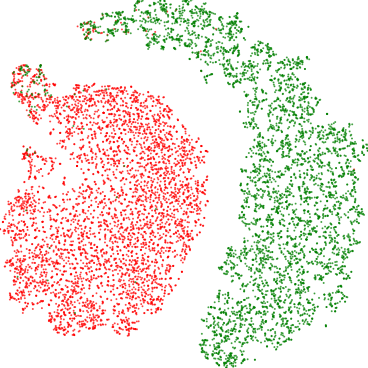}
      \end{minipage}
   }
   \end{center}
      \caption{The visualization of features supervised by (a) softmax loss, (b) triplet + softmax loss, (c) center + softmax loss, (d) SCL + softmax loss.
      We randomly selected 5000 natural faces and manipulated ones respectively from the training dataset of the FF++ c40 version. 
      Red dots represent natural faces and green dots represent manipulated faces (Best viewed in color). }
   \label{fig:visualization}
\end{figure*}

The influence of hyper-parameter $\lambda$ is presented in Figure~\ref{fig:para_1}. The experimental results show that
our SCL is quite robust to this parameter. For values from 0.001 to 1, 
the trained models consistently achieve promising results. We assume that is because SCL and softmax loss are complementary losses.
SCL focuses on feature representations directly, while softmax loss focuses on how to map feature representations into a discrete label space.
What's more, the global information retained by softmax loss can guide the update of the center point of SCL. 
When $\lambda$ is set to be 0, which means the model is trained by using only softmax loss, 
the performance is worst, only achieving an AUC of 0.861. But a $4\%\sim 6\%$ improvement of AUC could be reached by combining our SCL with softmax loss. 
In order to investigate the influence of m, we fix $\lambda$ to be 0.5, and then take seven values from 0.05 to 0.35 at 0.05 intervals as m.
It should be emphasized that m is a scale factor and the margin of the distance is proportional to the arithmetic square root of the dimension of the feature space, as shown in Eq.~\eqref{eq:SCL}.
As illustrated in Figure~\ref{fig:para_2}, our SCL can effectively improve the performance when m changes within a large range.
When m is set to be 0.3 and $\lambda$ to be 0.5, we get the best results, an AUC of 0.916 and a pAUC$_{0.1}$ of 0.790.
\paragraph{Comparison with other losses}
To validate the proposed SCL loss, we conduct additional experiments on various losses, 
including triplet loss with softmax loss and center loss with softmax loss.
Similar to Eq.~\eqref{eq:loss}, both the weight of triplet loss and center loss are set as 0.01 and the margin of triplet loss is set as 0.3.
As can be seen from Table~\ref{tab:losses}, our SCL loss with softmax loss performs best among these losses,
obtaining an AUC of 0.916 and a pAUC$_{0.1}$ of 0.790.
In addition, both triplet loss with softmax loss and center loss with softmax loss improve subtly, compared to only using softmax loss.
\begin{table}[h]
   \begin{center}      
\begin{tabular}{|l|c|c|}
  \hline
  Loss function&AUC&pAUC$_{0.1}$\\
  \hline
  softmax loss&0.860&0.652\\ 
  \hline
  center + softmax loss &0.868&0.666\\
  \hline
  triplet + softmax loss &0.863&0.655\\
  \hline
  SCL + softmax loss &\textbf{0.916}&\textbf{0.790}\\
  \hline
\end{tabular}
\end{center}
\caption{The performance of different losses on the c40 version of FF++.}
\label{tab:losses}
\vspace{-0.5cm}
\end{table}
\paragraph{Visualization of learned representations}
In order to explore the influence of different losses on feature distribution more thoroughly, 
we adopt t-SNE~\cite{maaten2008visualizing} to visualize features of the samples from the FF++ dataset.
As is shown in Figure~\ref{fig:visualization}, some properties can be observed: a) 
The learned features supervised by softmax loss appear as two clusters with neighboring boundaries.
b) The triplet loss has little effect on feature distribution. We have tried to increase the weight of triplet loss, but in such a case the network cannot converge normally.
c) The center loss significantly changes the distribution of features. However, constraining the intra-class compactness of manipulated faces leads to overfitting to some extent. 
Hence, the performance gain is very small.
d) With the combination of SCL + softmax loss, the representations of natural faces are gathered compactly and separated from those of manipulated faces which are distributed less compactly.
\paragraph{Results analysis}
As shown in Table~\ref{tab:losses} and Figure~\ref{fig:visualization}, our SCL outperforms other losses, $\ie$, softmax loss, center loss, and triplet loss.
It is no wonder that softmax loss performs poorly since it only focuses on finding a decision boundary to separate different classes.
As for triplet loss and center loss, though they explicitly consider intra-class compactness and inter-class separability,
the results show that indiscriminately constraining intra-class compactness of natural and manipulated faces usually leads to a sub-optimal solution.
This validates our analysis in Sec.~\ref{sec:SCL} that
different face manipulation methods will produce different forgery features due to GAN fingerprints ~\cite{yu2019attributing} and some unique operations, 
making it nontrivial to aggregate all of the manipulated faces together. 
Compared to them, our SCL adopts an asymmetric optimization goal for natural and manipulated faces to learn discriminative features, 
which is more compatible with the feature distribution of samples.
\subsubsection{Fusion module}
We have studied the effects of different structures of the fusion module on performance and all experiments are conducted only with the supervision of softmax loss.
As shown in Table~\ref{tab:fusion}, we explore concatenation, sum, and convolution block with different kernel sizes and group numbers. From the experimental results, We can see
that: a) When a $1\times 1$ convolution block is used as a fusion module and its group is set as 1, the performance reaches best in terms of AUC and pAUC$_{0.1}$;
b) Even simple concatenation and sum operation can still achieve good performance. 
This fully reflects the effectiveness of our adaptive frequency feature generation module. In order to achieve the best results, 
we utilize the $1\times 1$ convolution block, whose group is set to be 1, as the fusion module in all reference experiments.
\begin{table}[h]
   \begin{center}     
\begin{tabular}{|c|c|c|}
  \hline
   fusion module&AUC&pAUC$_{0.1}$\\
   \hline
   concatenation &0.892&0.727\\
  \hline
  sum &0.893&0.732\\ 
  \hline
  $3\times 3$, group=1 &0.894&0.708\\
  \hline
  $1\times 1$, group=1 &\textbf{0.906}&\textbf{0.769}\\
  \hline
  $3\times 3$, group=2 &0.899&0.735\\
  \hline
  $1\times 1$, group=2 &0.892&0.742\\
  \hline
\end{tabular}
\end{center}
\caption{The performance of different fusion modules on the c40 version of FF++.}
\label{tab:fusion}
\vspace{-0.2cm}
\end{table}
\begin{table}[h]
   \begin{center}     
\begin{tabular}{|cc|c|c|}
  \hline
  SCL&AFFGM&AUC&pAUC$_{0.1}$\\
  \hline
  -&-&0.861&0.652\\ 
  $\surd$&-&0.916&0.790\\
  -&$\surd$&0.906&0.769\\
  $\surd$&$\surd$&\textbf{0.924}&\textbf{0.810}\\
  \hline
\end{tabular}
\end{center}
\caption{The performance of different variants of FDFL on the c40 version of FF++.}
\label{tab:ablation}
\vspace{-0.4cm}
\end{table}
\begin{table*}
   \begin{center}
      \begin{tabular}{l|c|c|c|c|c|c|c|c|c}
         \hline
         \multirow{2}*{Methods} &\multicolumn{3}{|c|}{c0} &\multicolumn{3}{|c|}{c23}&\multicolumn{3}{|c}{c40}\\
         \cline{2-10} 
         &Acc&AUC&pAUC$_{0.1}$&Acc&AUC&pAUC$_{0.1}$&Acc&AUC&pAUC$_{0.1}$\\
         \hline
         Steg. Features + SVM\cite{fridrich2012rich} &97.63$\%$&-&-&70.97$\%$&-&-&55.98$\%$&-&-\\
         Cozzolino $\etal$\cite{cozzolino2017recasting} &98.57$\%$&-&-&78.45$\%$&-&-&58.69$\%$&-&-\\
         Bayar and Stamm\cite{bayar2016deep} &98.75$\%$&-&-&82.97$\%$&-&-&66.84$\%$&-&-\\
         Rahmouni $\etal$\cite{rahmouni2017distinguishing} &97.03$\%$&-&-&79.08$\%$&-&-&61.18$\%$&-&-\\
         DSP-FWA\cite{li2018exposing} &-&-&-&-&0.575&0.516&-&0.623&0.519\\
         MesoNet\cite{afchar2018mesonet} &95.23$\%$&-&-&83.10$\%$&-&-&70.47$\%$&-&-\\
         Xception\cite{rossler2019faceforensics++} &99.26$\%$&-&-&95.73$\%$&-&-&81.00$\%$&-&-\\
         Face X-ray\cite{li2020face} &-&0.988&-&-&0.874&-&-&0.616&-\\
         \hline
         Two-branch\cite{masi2020two} &-&-&-&96.43$\%$&0.991&0.984&86.34$\%$&0.911&0.766\\
         \hline
         Xception\cite{rossler2019faceforensics++} $\dag$ &98.14$\%$&0.997$^{*}$&0.995&94.24$\%$&0.972&0.903&86.14$\%$&0.861&0.652\\
         FDFL(our) &\textbf{99.43$\%$}&\textbf{0.997}$^{*}$&\textbf{0.998}&\textbf{96.69}$\%$&\textbf{0.993}&\textbf{0.985}&\textbf{89.00$\%$}&\textbf{0.924}&\textbf{0.810}\\
       \hline 
       \end{tabular}
       \end{center}
       \caption{Quantitative results on the FF++ dataset with all three versions.
       c0 represents videos without compression, c23 represents videos with light compression, c40 represents videos with heavy compression and $\dag$ represents the results of our baseline.
       Two-branch~\cite{masi2020two} is a video-based detection method and all others are image-based detection methods.
       The bold results are the best. The symbol * represents there is a difference at the fourth decimal place and
       more precise data are provided in the supplementary material.}
       \label{tab:methods}
       \vspace{-0.4cm}
       \end{table*}
\subsubsection{The performance gain of each component}
In order to evaluate the performance improvement from each component, we quantitatively evaluate our FDFL framework and its variants: 1) the baseline (xception); 2) FDFL w/o SCL;
3) FDFL w/o AFFGM. The quantitative results are listed in Table~\ref{tab:ablation}. It can be seen that both SCL and AFFGM can boost performance in terms of AUC and pAUC$_{0.1}$.
Specifically, only with our SCL, AUC and pAUC$_{0.1}$ increased to 0.916 and 0.79 with an improvement of 5.5$\%$ and 13.8$\%$ individually. And AFFGM can also contribute an improvement of $4.5\%$ in terms
of AUC and $11.7\%$ in terms of pAUC$_{0.1}$. These improvements prove the effectiveness of our SCL and AFFGM. In addition, when SCL and AFFGM are simultaneously integrated into the baseline to form the complete FDFL framework, an AUC of 0.924 and a pAUC$_{0.1}$ of 0.810 can be obtained.
This fully validates the ability of our SCL to supervise the network to learn more discriminative features.
\subsection{Comparison with previous methods}
We compare our approach with previous face forgery detection methods on the FF++ dataset.
The results are listed in Table~\ref{tab:methods}. Xception~\cite{rossler2019faceforensics++} and Face X-ray~\cite{li2020face}
are currently state-of-the-art image-based detection methods. Our method outperforms them on various versions of FF++ dataset 
in terms of AUC, pAUC$_{0.1}$, and accuracy. In the most challenging c40 version, we achieve a 6.4$\%$, 15.8$\%$, and 2.86$\%$ improvement in AUC, pAUC$_{0.1}$, and accuracy respectively.
Although two-branch~\cite{masi2020two} is a video-based detection method, we still 
surpass it, especially in the c40 version.
The results demonstrate the effectiveness and superiority of our framework.
\section{Limitations}
Although we achieve remarkable results on the FF++ dataset, there exist limitations of our framework.
On the one hand, our framework lacks generalization ability on unseen manipulation methods (the results are provided in supplementary material). Our work and Face X-ray~\cite{li2020face} imply that 
the discriminative features learned by supervised learning are highly related to manipulation methods.
In our view, that is because forgery evidence is customized for specific manipulation methods due to GAN fingerprints~\cite{yu2019attributing} and some unique operations. 
If this explanation holds, not only our approach, but all approaches based on supervised learning will lack generalization ability on unseen manipulation methods.
On the other hand, our framework ignores inter-frame information. Current face manipulation methods generally do not impose constraints on the temporal dimension.
Therefore, the inconsistency between frames should be a valuable cue for video face forgery detection. 
\section{Conclusion}
In this paper, we propose a novel frequency-aware discriminative feature learning framework that applies metric learning and 
adaptive frequency features learning to face forgery detection.
Specifically, our single-center loss only compresses intra-class variations of natural faces when boosting inter-class separability in the embedding space. 
In such a case, the network can learn more discriminative features with less optimization difficulty.
Besides, our adaptive frequency features generation module can effectively mine subtle artifacts from the frequency domain in a data-driven fashion, which avoids overly
depending on incomprehensive prior knowledge.
Extensive experiments demonstrate the effectiveness and superiority of our FDFL and we achieve state-of-the-art results on three versions of FF++ dataset.

In the future, we will explore how to effectively exploit inter-frame information and improve the generalization ability of detection methods by semi-supervised and unsupervised learning.
In addition, it is worth studying the application of SCL in other fields, such as face anti-spoofing.
\vspace{-0.2cm}
\paragraph{Acknowledgements}
This work is supported by the National Key Research and Development Program of China 
(2017YFC0820600), the National Nature Science Foundation of China (62022076, U1936210), 
the Youth Innovation Promotion Association Chinese Academy of Sciences (2017209).
{\small
\bibliographystyle{ieee_fullname}
\bibliography{egbib}

\begin{thebibliography}{10}\itemsep=-1pt

\bibitem{Deepfake}
Deepfakes.
\newblock \url{https://www.github.com/deepfakes/faceswap}.
\newblock Accessed: 2019-09-18.

\bibitem{FaceApp}
Faceapp.
\newblock \url{http://faceapp.com/app}.
\newblock Accessed: 2019-09-04.

\bibitem{FaceSwap01}
Faceswap.
\newblock \url{https://www.github.com/MarekKowalski/FaceSwap}.
\newblock Accessed: 2019-09-30.

\bibitem{afchar2018mesonet}
Darius Afchar, Vincent Nozick, Junichi Yamagishi, and Isao Echizen.
\newblock Mesonet: a compact facial video forgery detection network.
\newblock In {\em 2018 IEEE International Workshop on Information Forensics and
  Security (WIFS)}, pages 1--7. IEEE, 2018.

\bibitem{agarwal2019protecting}
Shruti Agarwal, Hany Farid, Yuming Gu, Mingming He, Koki Nagano, and Hao Li.
\newblock Protecting world leaders against deep fakes.
\newblock In {\em CVPR Workshops}, pages 38--45, 2019.

\bibitem{bayar2016deep}
Belhassen Bayar and Matthew~C Stamm.
\newblock A deep learning approach to universal image manipulation detection
  using a new convolutional layer.
\newblock In {\em Proceedings of the 4th ACM Workshop on Information Hiding and
  Multimedia Security}, pages 5--10, 2016.

\bibitem{chollet2017xception}
Fran{\c{c}}ois Chollet.
\newblock Xception: Deep learning with depthwise separable convolutions.
\newblock In {\em Proceedings of the IEEE conference on computer vision and
  pattern recognition}, pages 1251--1258, 2017.

\bibitem{cozzolino2017recasting}
Davide Cozzolino, Giovanni Poggi, and Luisa Verdoliva.
\newblock Recasting residual-based local descriptors as convolutional neural
  networks: an application to image forgery detection.
\newblock In {\em Proceedings of the 5th ACM Workshop on Information Hiding and
  Multimedia Security}, pages 159--164, 2017.

\bibitem{cozzolino2018forensictransfer}
Davide Cozzolino, Justus Thies, Andreas R{\"o}ssler, Christian Riess, Matthias
  Nie{\ss}ner, and Luisa Verdoliva.
\newblock Forensictransfer: Weakly-supervised domain adaptation for forgery
  detection.
\newblock {\em arXiv preprint arXiv:1812.02510}, 2018.

\bibitem{dang2020detection}
Hao Dang, Feng Liu, Joel Stehouwer, Xiaoming Liu, and Anil~K Jain.
\newblock On the detection of digital face manipulation.
\newblock In {\em Proceedings of the IEEE/CVF Conference on Computer Vision and
  Pattern Recognition}, pages 5781--5790, 2020.

\bibitem{deng2020retinaface}
Jiankang Deng, Jia Guo, Evangelos Ververas, Irene Kotsia, and Stefanos
  Zafeiriou.
\newblock Retinaface: Single-shot multi-level face localisation in the wild.
\newblock In {\em Proceedings of the IEEE/CVF Conference on Computer Vision and
  Pattern Recognition}, pages 5203--5212, 2020.

\bibitem{deng2020disentangled}
Yu Deng, Jiaolong Yang, Dong Chen, Fang Wen, and Xin Tong.
\newblock Disentangled and controllable face image generation via 3d
  imitative-contrastive learning.
\newblock In {\em Proceedings of the IEEE/CVF Conference on Computer Vision and
  Pattern Recognition}, pages 5154--5163, 2020.

\bibitem{durall2019unmasking}
Ricard Durall, Margret Keuper, Franz-Josef Pfreundt, and Janis Keuper.
\newblock Unmasking deepfakes with simple features.
\newblock {\em arXiv preprint arXiv:1911.00686}, 2019.

\bibitem{fridrich2012rich}
Jessica Fridrich and Jan Kodovsky.
\newblock Rich models for steganalysis of digital images.
\newblock {\em IEEE Transactions on Information Forensics and Security},
  7(3):868--882, 2012.

\bibitem{goodfellow2014generative}
Ian Goodfellow, Jean Pouget-Abadie, Mehdi Mirza, Bing Xu, David Warde-Farley,
  Sherjil Ozair, Aaron Courville, and Yoshua Bengio.
\newblock Generative adversarial nets.
\newblock In {\em Advances in neural information processing systems}, pages
  2672--2680, 2014.

\bibitem{gueguen2018faster}
Lionel Gueguen, Alex Sergeev, Ben Kadlec, Rosanne Liu, and Jason Yosinski.
\newblock Faster neural networks straight from jpeg.
\newblock In {\em Advances in Neural Information Processing Systems}, pages
  3933--3944, 2018.

\bibitem{hermans2017defense}
Alexander Hermans, Lucas Beyer, and Bastian Leibe.
\newblock In defense of the triplet loss for person re-identification.
\newblock {\em arXiv preprint arXiv:1703.07737}, 2017.

\bibitem{ioannou2017deep}
Yani Ioannou, Duncan Robertson, Roberto Cipolla, and Antonio Criminisi.
\newblock Deep roots: Improving cnn efficiency with hierarchical filter groups.
\newblock In {\em Proceedings of the IEEE conference on computer vision and
  pattern recognition}, pages 1231--1240, 2017.

\bibitem{jung2020deepvision}
Tackhyun Jung, Sangwon Kim, and Keecheon Kim.
\newblock Deepvision: Deepfakes detection using human eye blinking pattern.
\newblock {\em IEEE Access}, 8:83144--83154, 2020.

\bibitem{karras2019style}
Tero Karras, Samuli Laine, and Timo Aila.
\newblock A style-based generator architecture for generative adversarial
  networks.
\newblock In {\em Proceedings of the IEEE conference on computer vision and
  pattern recognition}, pages 4401--4410, 2019.

\bibitem{kingma2013auto}
Diederik~P Kingma and Max Welling.
\newblock Auto-encoding variational bayes.
\newblock {\em arXiv preprint arXiv:1312.6114}, 2013.

\bibitem{kumar2020detecting}
Akash Kumar, Arnav Bhavsar, and Rajesh Verma.
\newblock Detecting deepfakes with metric learning.
\newblock In {\em 2020 8th International Workshop on Biometrics and Forensics
  (IWBF)}, pages 1--6. IEEE, 2020.

\bibitem{lee2020maskgan}
Cheng-Han Lee, Ziwei Liu, Lingyun Wu, and Ping Luo.
\newblock Maskgan: Towards diverse and interactive facial image manipulation.
\newblock In {\em Proceedings of the IEEE/CVF Conference on Computer Vision and
  Pattern Recognition}, pages 5549--5558, 2020.

\bibitem{li2020face}
Lingzhi Li, Jianmin Bao, Ting Zhang, Hao Yang, Dong Chen, Fang Wen, and Baining
  Guo.
\newblock Face x-ray for more general face forgery detection.
\newblock In {\em Proceedings of the IEEE/CVF Conference on Computer Vision and
  Pattern Recognition}, pages 5001--5010, 2020.

\bibitem{li2018exposing}
Yuezun Li and Siwei Lyu.
\newblock Exposing deepfake videos by detecting face warping artifacts.
\newblock {\em arXiv preprint arXiv:1811.00656}, 2018.

\bibitem{liu2019bidirectional}
Chuanbin Liu, Hongtao Xie, Zhengjun Zha, Lingyun Yu, Zhineng Chen, and Yongdong
  Zhang.
\newblock Bidirectional attention-recognition model for fine-grained object
  classification.
\newblock {\em IEEE Transactions on Multimedia}, 22(7):1785--1795, 2019.

\bibitem{liu2016large}
Weiyang Liu, Yandong Wen, Zhiding Yu, and Meng Yang.
\newblock Large-margin softmax loss for convolutional neural networks.
\newblock In {\em ICML}, volume~2, page~7, 2016.

\bibitem{maaten2008visualizing}
Laurens van~der Maaten and Geoffrey Hinton.
\newblock Visualizing data using t-sne.
\newblock {\em Journal of machine learning research}, 9(Nov):2579--2605, 2008.

\bibitem{masi2020two}
Iacopo Masi, Aditya Killekar, Royston~Marian Mascarenhas, Shenoy~Pratik
  Gurudatt, and Wael AbdAlmageed.
\newblock Two-branch recurrent network for isolating deepfakes in videos.
\newblock {\em arXiv preprint arXiv:2008.03412}, 2020.

\bibitem{mcclish1989analyzing}
Donna~Katzman McClish.
\newblock Analyzing a portion of the roc curve.
\newblock {\em Medical Decision Making}, 9(3):190--195, 1989.

\bibitem{min2020multi}
Shaobo Min, Hantao Yao, Hongtao Xie, Zheng-Jun Zha, and Yongdong Zhang.
\newblock Multi-objective matrix normalization for fine-grained visual
  recognition.
\newblock {\em IEEE Transactions on Image Processing}, 29:4996--5009, 2020.

\bibitem{paszke2019pytorch}
Adam Paszke, Sam Gross, Francisco Massa, Adam Lerer, James Bradbury, Gregory
  Chanan, Trevor Killeen, Zeming Lin, Natalia Gimelshein, Luca Antiga, et~al.
\newblock Pytorch: An imperative style, high-performance deep learning library.
\newblock In {\em Advances in neural information processing systems}, pages
  8026--8037, 2019.

\bibitem{qi2020deeprhythm}
Hua Qi, Qing Guo, Felix Juefei-Xu, Xiaofei Xie, Lei Ma, Wei Feng, Yang Liu, and
  Jianjun Zhao.
\newblock Deeprhythm: Exposing deepfakes with attentional visual heartbeat
  rhythms.
\newblock In {\em Proceedings of the 28th ACM International Conference on
  Multimedia}, pages 4318--4327, 2020.

\bibitem{rahmouni2017distinguishing}
Nicolas Rahmouni, Vincent Nozick, Junichi Yamagishi, and Isao Echizen.
\newblock Distinguishing computer graphics from natural images using
  convolution neural networks.
\newblock In {\em 2017 IEEE Workshop on Information Forensics and Security
  (WIFS)}, pages 1--6. IEEE, 2017.

\bibitem{rezende2014stochastic}
Danilo~Jimenez Rezende, Shakir Mohamed, and Daan Wierstra.
\newblock Stochastic backpropagation and approximate inference in deep
  generative models.
\newblock {\em arXiv preprint arXiv:1401.4082}, 2014.

\bibitem{rossler2019faceforensics++}
Andreas Rossler, Davide Cozzolino, Luisa Verdoliva, Christian Riess, Justus
  Thies, and Matthias Nie{\ss}ner.
\newblock Faceforensics++: Learning to detect manipulated facial images.
\newblock In {\em Proceedings of the IEEE International Conference on Computer
  Vision}, pages 1--11, 2019.

\bibitem{schroff2015facenet}
Florian Schroff, Dmitry Kalenichenko, and James Philbin.
\newblock Facenet: A unified embedding for face recognition and clustering.
\newblock In {\em Proceedings of the IEEE conference on computer vision and
  pattern recognition}, pages 815--823, 2015.

\bibitem{shen2020interpreting}
Yujun Shen, Jinjin Gu, Xiaoou Tang, and Bolei Zhou.
\newblock Interpreting the latent space of gans for semantic face editing.
\newblock In {\em Proceedings of the IEEE/CVF Conference on Computer Vision and
  Pattern Recognition}, pages 9243--9252, 2020.

\bibitem{thies2019deferred}
Justus Thies, Michael Zollh{\"o}fer, and Matthias Nie{\ss}ner.
\newblock Deferred neural rendering: Image synthesis using neural textures.
\newblock {\em ACM Transactions on Graphics (TOG)}, 38(4):1--12, 2019.

\bibitem{thies2016face2face}
Justus Thies, Michael Zollhofer, Marc Stamminger, Christian Theobalt, and
  Matthias Nie{\ss}ner.
\newblock Face2face: Real-time face capture and reenactment of rgb videos.
\newblock In {\em Proceedings of the IEEE conference on computer vision and
  pattern recognition}, pages 2387--2395, 2016.

\bibitem{wang2020cnn}
Sheng-Yu Wang, Oliver Wang, Richard Zhang, Andrew Owens, and Alexei~A Efros.
\newblock Cnn-generated images are surprisingly easy to spot... for now.
\newblock In {\em Proceedings of the IEEE Conference on Computer Vision and
  Pattern Recognition}, volume~7, 2020.

\bibitem{wen2016discriminative}
Yandong Wen, Kaipeng Zhang, Zhifeng Li, and Yu Qiao.
\newblock A discriminative feature learning approach for deep face recognition.
\newblock In {\em European conference on computer vision}, pages 499--515.
  Springer, 2016.

\bibitem{wu2020cascade}
Rongliang Wu, Gongjie Zhang, Shijian Lu, and Tao Chen.
\newblock Cascade ef-gan: Progressive facial expression editing with local
  focuses.
\newblock In {\em Proceedings of the IEEE/CVF Conference on Computer Vision and
  Pattern Recognition}, pages 5021--5030, 2020.

\bibitem{Xu_2020_CVPR}
Kai Xu, Minghai Qin, Fei Sun, Yuhao Wang, Yen-Kuang Chen, and Fengbo Ren.
\newblock Learning in the frequency domain.
\newblock In {\em IEEE/CVF Conference on Computer Vision and Pattern
  Recognition (CVPR)}, June 2020.

\bibitem{yang2020one}
Chao Yang and Ser-Nam Lim.
\newblock One-shot domain adaptation for face generation.
\newblock In {\em Proceedings of the IEEE/CVF Conference on Computer Vision and
  Pattern Recognition}, pages 5921--5930, 2020.

\bibitem{yang2019exposing}
Xin Yang, Yuezun Li, and Siwei Lyu.
\newblock Exposing deep fakes using inconsistent head poses.
\newblock In {\em ICASSP 2019-2019 IEEE International Conference on Acoustics,
  Speech and Signal Processing (ICASSP)}, pages 8261--8265. IEEE, 2019.

\bibitem{yu2019attributing}
Ning Yu, Larry~S Davis, and Mario Fritz.
\newblock Attributing fake images to gans: Learning and analyzing gan
  fingerprints.
\newblock In {\em Proceedings of the IEEE International Conference on Computer
  Vision}, pages 7556--7566, 2019.

\end{thebibliography}
}

\end{document}